\documentclass[11pt,letterpaper]{article}
\usepackage[letterpaper]{geometry}
\usepackage{acl2012}
\usepackage{times}
\usepackage{latexsym}
\usepackage{amsmath}
\usepackage{mathtools}
\usepackage{amssymb}
\usepackage{multirow}
\usepackage{graphicx}
\usepackage{todonotes}
\usepackage{url}
\usepackage{color}
\usepackage[inline]{enumitem}
\usepackage{booktabs}
\usepackage[font=small]{caption}
\usepackage[font=small]{subcaption}
\usepackage{colortbl}
\usepackage{tikz}
\usepackage{comment}
\usepackage{epstopdf}
\usetikzlibrary{shapes.geometric}
\makeatletter
\newcommand{\@BIBLABEL}{\@emptybiblabel}
\newcommand{\@emptybiblabel}[1]{}
\makeatother
\usepackage[hidelinks]{hyperref}

\makeatletter

\def\citealt{\def\citename##1{{\frenchspacing##1} }\@internalcitec}

\def\@citexc[#1]#2{\if@filesw\immediate\write\@auxout{\string\citation{#2}}\fi
  \def\@citea{}\@citealt{\@for\@citeb:=#2\do
    {\@citea\def\@citea{;\penalty\@m\ }\@ifundefined
       {b@\@citeb}{{\bf ?}\@warning
       {Citation `\@citeb' on page \thepage \space undefined}}%
{\csname b@\@citeb\endcsname}}}{#1}}

\def\@internalcitec{\@ifnextchar [{\@tempswatrue\@citexc}{\@tempswafalse\@citexc[]}}

\def\@citealt#1#2{{#1\if@tempswa, #2\fi}}

\makeatother

\newcommand{\term}{\textit}

\newcommand{\word}{\textit}

\newcommand{\eq}[1]{(\ref{#1})}

\newcommand{\Listener}{L}
\newcommand{\Speaker}{S}
\newcommand{\utt}{u}

\newcommand{\referent}{c}
\newcommand{\context}{C}
\newcommand{\contextlen}{K}
\newcommand{\target}{t}
\newcommand{\numsamples}{m}
\newcommand{\feat}{f}
\renewcommand{\|}{\mid}
\newcommand{\best}[1]{\textbf{#1}}
\newcommand{\oracle}[1]{\textit{#1}}

\newcommand{\set}[1]{\left\{#1\right\}}

\newcommand{\secref}[1]{Section~\ref{#1}}

\newcommand{\Figref}[1]{Figure~\ref{#1}}
\newcommand{\figref}[1]{Figure~\ref{#1}}

\newcommand{\Tabref}[1]{Table~\ref{#1}}
\newcommand{\tabref}[1]{Table~\ref{#1}}

\definecolor{ourlightblue}{HTML}{03A9F4}
\definecolor{ourgreen}{HTML}{4D8951}
\definecolor{oursteelblue}{HTML}{9BB8D7}
\definecolor{ourorange}{HTML}{FDBA58}

\newcolumntype{P}[1]{>{\raggedright\arraybackslash}p{#1}}

\newcommand{\colorPatch}[2][xxxx]{
  \colorbox[HTML]{#2}{{\color[HTML]{#2}#1}}}

\newcommand{\colorContext}[4]{
  \framebox{\negthickspace\colorPatch{#1}} & \colorPatch{#2} & \colorPatch{#3} & #4}

\newcommand{\colorContextCompact}[3]{
  \colorPatch[xx]{#1} & \colorPatch[xx]{#2} & \colorPatch[xx]{#3}}

\newcommand{\colorContextNarrow}[3]{
  \colorPatch[xx]{#1}, \colorPatch[xx]{#2}, \colorPatch[xx]{#3}}

\newcommand{\gzz}{\phantom{$<$00}}
\newcommand{\gz}{\phantom{$<$0}}
\newcommand{\zz}{\phantom{00}}
\newcommand{\z}{\phantom{0}}
\newcommand{\p}{}

 \definecolor{Green}{RGB}{10,200,100}

\newcommand{\cond}{\emph}

\newcommand{\email}{\texttt}
\newcommand\authmark[1]{\textnormal{\textsuperscript{#1}}}

\title{Colors in Context: A Pragmatic Neural Model for \\
Grounded Language Understanding}

\author{
Will Monroe,\authmark{1}\;
Robert X.D. Hawkins,\authmark{2}\;
Noah D. Goodman,\authmark{1,2}
\and Christopher Potts\authmark{3} \\
Departments of \authmark{1}Computer Science, \authmark{2}Psychology, and \authmark{3}Linguistics \\
Stanford University, Stanford, CA 94305 \\
\email{wmonroe4@cs.stanford.edu}, \{\email{rxdh}, \email{ngoodman}, \email{cgpotts}\}\email{@stanford.edu}
}

\date{}

\begin{document}
\maketitle
\begin{abstract}

We present a model of pragmatic referring expression interpretation
 in a grounded communication task
(identifying colors from descriptions) that draws upon predictions
from two recurrent neural network classifiers, a speaker and a listener,
unified by a recursive pragmatic reasoning framework.
Experiments show that this
combined pragmatic model interprets color descriptions
more accurately than the classifiers from which it is built, and that
much of this improvement results from combining the speaker and listener
perspectives.
We observe that pragmatic reasoning helps primarily
in the hardest cases: when the model must distinguish very similar colors,
or when few utterances adequately express
the target color.
Our findings make use of a newly-collected corpus of human utterances in
color reference games, which exhibit a variety of pragmatic behaviors.
We also show that the embedded speaker model reproduces many of these
 pragmatic behaviors.

\end{abstract}

\section{Introduction} \label{sec:intro}

Human communication is \emph{situated}. In using language, we are sensitive
to context and our interlocutors' expectations, both when choosing our
utterances (as speakers) and when interpreting the utterances we hear (as listeners).
Visual referring tasks exercise this complex process of grounding,
in the environment and in
our mental models of each other, and thus provide a valuable test-bed for
computational models of production and comprehension.

\begin{table}
  \centering
  \setlength{\tabcolsep}{4pt}
  \begin{tabular}[c]{r@{. \ } ccc l}
    \toprule
    \multicolumn{4}{c}{Context} & Utterance \\
    \midrule
    1&\colorContext{2421DE}{605DA2}{0144FE}{darker blue}\\
    2&\colorContext{5866A7}{2DD2BC}{C23D5A}{Purple}\\
    3&\colorContext{5866A7}{9953AC}{2DD2A6}{blue}\\
    4&\colorContext{3884C7}{02F9FD}{9E6461}{blue}\\
    \bottomrule
  \end{tabular}
  \caption{Examples of color reference in context, taken from our corpus. The target color
    is boxed. The speaker's description is shaped not only
    by this target, but also by the other context colors and their
    relationships.}
  \label{table:examples}
\end{table}

\Tabref{table:examples} illustrates the situated nature of
reference understanding with descriptions of colors from a task-oriented
dialogue corpus we
introduce in this paper. In these dialogues, the speaker is trying
to identify their (privately assigned) target color for the
listener. In context~1, the comparative \word{darker} implicitly
refers to both the target (boxed) and one of the other colors. In
contexts 2 and 3, the target color is the same, but the distractors
led the speaker to choose different basic color terms. In
context~4, \word{blue} is a pragmatic choice even though two colors are
shades of blue, because the interlocutors assume about each other that
they find the target color a more prototypical representative of blue
and would prefer other descriptions (\word{teal}, \word{cyan}) for the middle color.
The fact that \word{blue} appears in three of these
four cases highlights the flexibility and context dependence of color descriptions.

In this paper, we present a scalable, learned model of pragmatic
language understanding. The model is built around a version of the
Rational Speech Acts (RSA) model \cite{Frank2012,GoodmanFrank16_RSATiCS}, in which agents
reason recursively about each other's expectations and intentions to
communicate more effectively than literal semantic agents could. In
most work on RSA, the literal semantic agents use fixed message sets
and stipulated grammars, which is a barrier to experiments in
linguistically complex domains. In our formulation, the literal
semantic agents are recurrent neural networks (RNNs) that produce and
interpret color descriptions in context. These models are learned from
data and scale easily to large datasets containing diverse utterances.
The RSA recursion is then defined in terms of these base agents: the
\emph{pragmatic speaker} produces utterances based on a literal RNN
listener \cite{AndreasKlein16_NeuralPragmatics}, and the
\emph{pragmatic listener} interprets utterances based on the pragmatic
speaker's behavior.

We focus on accuracy in a listener task
(i.e., at language understanding).
However, our most successful model integrates speaker and listener perspectives,
combining predictions made by a system trained to understand color descriptions
and one trained to produce them.

We evaluate this model with a new, psycholinguistically motivated corpus of real-time, dyadic reference games in which
the referents are patches of color.
Our task is fundamentally the same as that of
\newcite{Baumgaertner2012}, but the corpus we release is larger by several
orders of magnitude, consisting of 948 complete games with 53,365 utterances 
produced by human participants paired into dyads on the web. The linguistic
behavior of the players exhibits many of the intricacies of language
in general, including not just the context dependence and cognitive
complexity discussed above, but also compositionality, vagueness, and
ambiguity. While many previous data sets feature descriptions of
individual colors \cite{Cook2005,Munroe2010,Kawakami2016}, situating
colors in a communicative context elicits greater variety in language use,
including negations, comparatives, superlatives,
metaphor, and shared associations.

Experiments on the data in our corpus show that this combined pragmatic model
improves accuracy
in interpreting human-produced descriptions over the basic RNN listener
alone. We find that the largest improvement over the single RNN comes from
blending it with an RNN trained to perform the speaker task, despite the fact
that a model based only on this speaker RNN performs poorly on its own.
Pragmatic reasoning on top of the listener RNN alone also yields improvements,
which moreover come primarily in the hardest cases:
\begin{enumerate*}[label=\arabic*)]
\item contexts with colors that are very similar, thus requiring
  the interpretation of descriptions that convey fine distinctions;
  and
\item target colors that most referring expressions fail to identify,
  whether due to a lack of adequate descriptive terms or a consistent
  bias against the color in the RNN listener.
\end{enumerate*}

\section{Task and data collection}\label{sec:corpus}

We evaluate our agents on a task of language understanding in a dyadic reference
game \cite{Rosenberg:Cohen:1964,KraussWeinheimer64_ReferencePhrases,Paetzel-etal:2014}. Unlike traditional natural language processing tasks, in which participants provide impartial judgements of language in isolation, reference games embed language use in a goal-oriented communicative context \cite{Clark96,TanenhausBrownSchmidt08_LanguageNatural}. Since they offer the simplest experimental setup where many pragmatic and discourse-level phenomena emerge, these games have been used widely in cognitive science to study topics like common ground and conventionalization \cite{Clark:Wilkes-Gibbs:1986}, referential domains \cite{BrownSchmidtTanenhaus08_TargetedGame}, perspective-taking \cite{HannaTanenhausTrueswell03_CommonGroundPerspective}, and overinformativeness \cite{KoolenGattGoudbeekKrahmer11_Overspecification}.

\begin{figure}
\includegraphics[scale = .2]{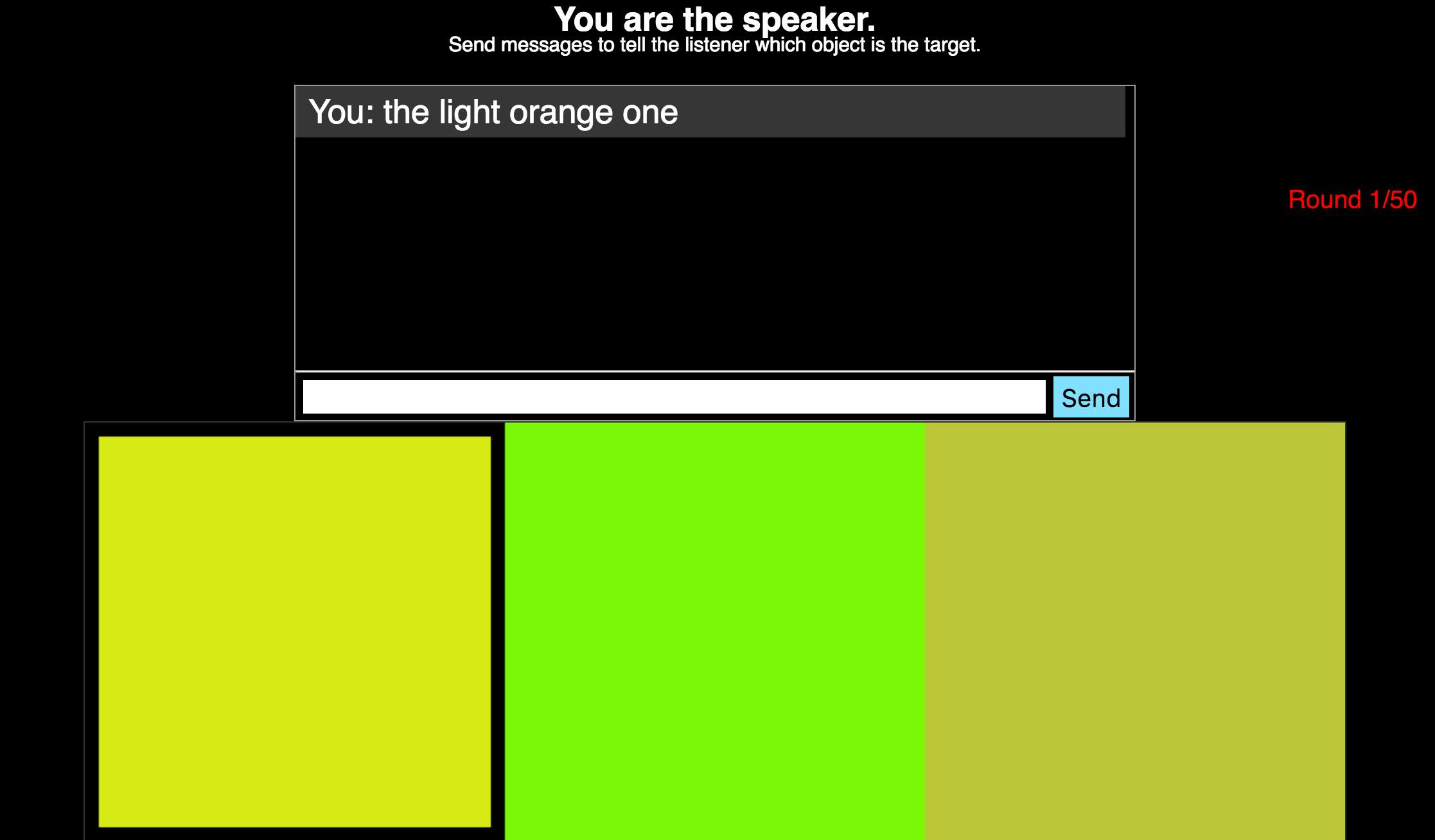}
\caption{Example trial in corpus collection task, from speaker's
  perspective. The target color (boxed) was presented among two distractors on a neutral background.}
\label{fig:taskScreenshot}
\end{figure}
To obtain a corpus of natural color reference data across varying
contexts, we recruited 967 unique participants from Amazon
Mechanical Turk to play 1,059 games of 50 rounds each, using the open-source framework of
\newcite{Hawkins15_RealTimeWebExperiments}.
Participants were sorted into dyads, randomly assigned the role of speaker or listener,
and placed in a game environment containing a chat box and an array of three color patches
(\figref{fig:taskScreenshot}).
On each round, one of the three colors was chosen to be the target and
highlighted for the speaker. They were instructed to communicate this
information to the listener, who could then click on one of the colors
to advance to the next trial. Both participants were free to use the
chat box at any point.

To ensure a range of difficulty, we randomly interspersed an equal
number of trials from three different conditions:
\begin{enumerate*}[label=\arabic*)]%
\item \cond{close}, where colors were all within a distance of
  $\theta$ from one another but still perceptible,\footnote{We used the
    most recent CIEDE standard to measure color differences, which is
    calibrated to human vision \cite{SharmaWuDalal05_DeltaE}. All
    distances were constrained to be larger than a lower bound of
    $\epsilon = 5$ to ensure perceptible differences, and we used a
    threshold value of $\theta = 20$ to create conditions.}
\item \cond{split}, where one distractor was within a distance of
  $\theta$ of the target, but the other distractor was farther than
  $\theta$, and
\item \cond{far}, where all colors were farther than $\theta$ from one
  another. Colors were rejection sampled uniformly from RGB (red,
  green, blue) space to meet these constraints.
\end{enumerate*}

After excluding extremely long messages,\footnote{Specifically, we set a length criterion at $4\sigma$ of the mean number of words per message (about 14 words, in our case), excluding 627 utterances. These often included meta-commentary about the game rather than color terms.} incomplete games, and games whose participants self-reported confusion about the instructions or non-native English proficiency, we were left with a corpus of 53,365 speaker utterances across 46,994 rounds in 948 games. The three conditions are equally represented, with 15,519 \emph{close} trials, 15,693 \emph{split} trials, and 15,782 \emph{far} trials. Participants were allowed to play more than once, but the modal number of games played per participant was one (75\%). The modal number of messages sent per round was also one (90\%). We release the filtered corpus we used throughout our analyses alongside the raw, pre-filter data collected from these experiments (see Footnote
\ref{foot:release}).

\section{Behavioral results}

Our corpus was developed not only to facilitate the development of
models for grounded language understanding, but also to provide a
richer picture of human pragmatic communication. The collection effort
was thus structured like a large-scale behavioral experiment, closely
following experimental designs like those of
\newcite{Clark:Wilkes-Gibbs:1986}. This paves the way to assessing our
model not solely based on the listener's classification accuracy, but also in terms of how
qualitative features of the speaker's production compare to that of our human participants.
Thus, the current section briefly reviews some novel findings from the human
corpus that we use to inform our model assessment.

\subsection{Listener behavior}

Since color reference is a difficult task even for humans, we compared listener accuracy across conditions to calibrate our expectations about model performance. While participants' accuracy was close to ceiling (97\%) on the \cond{far} condition, they made significantly more errors on the \cond{split} (90\%) and \cond{close} (83\%) conditions (see \figref{fig:listenerAccuracy}).

\begin{table*}[ht]
\centering
\renewcommand{\arraystretch}{0.9}
\begin{tabular}{lrrrr@{\hspace{20pt}}rrrr@{\hspace{20pt}}rrr}
  \toprule
  & \multicolumn{3}{c}{human}&& \multicolumn{3}{c}{$\Speaker_0$}&& \multicolumn{3}{c}{$\Speaker_1$}\\
  & far& split& close&& far& split& close&& far& split& close\\ \midrule
  \# Chars            & 7.8 & 12.3 & 14.9 && 9.0 & 12.8 & 16.6 && 9.0 & 12.8 & 16.4 \\
  \# Words            & 1.7 & 2.7 & 3.3   && 2.0 & 2.8 & 3.7   && 2.0 & 2.8 & 3.7   \\
  \% Comparatives     & 1.7 & 14.2 & 12.8 && 3.6 & 8.8 & 13.1  && 4.2 & 9.0 & 13.7  \\
  \% High Specificity & 7.0 & 7.6 & 7.4   && 6.4 & 8.4 & 7.6   && 6.8 & 7.9 & 7.5   \\
  \% Negatives        & 2.8 & 10.0 & 12.9 && 4.8 & 8.9 & 13.3  && 4.4 & 8.5 & 14.1  \\
  \% Superlatives     & 2.2 & 6.1 & 16.7  && 4.7 & 9.7 & 17.2  && 4.8 & 10.3 & 16.6 \\
   \bottomrule
\end{tabular}
\caption{Corpus statistics and statistics of samples from artificial speakers
  (rates per utterance). $\Speaker_0$: RNN speaker; $\Speaker_1$: pragmatic speaker 
  derived from RNN listener (see \secref{sec:l2}). The human and artificial speakers
  show many of the same correlations between language use and context type.}
\label{table:metrics}
\end{table*}

\subsection{Speaker behavior} \label{sec:speaker_behavior}

For ease of comparison to computational results, we focus on five
metrics capturing different aspects of pragmatic behavior
displayed by both human and artificial speakers in our task
(\tabref{table:metrics}). In all cases, we report test statistics from a mixed-effects regression including condition as a fixed effect and game ID as a random effect; except where noted, all test statistics reported correspond to $p$-values${}<10^{-4}$ and have been omitted for readability.

\paragraph{Words and characters}
We expect human speakers to be more verbose in \cond{split} and \cond{close}
contexts than \cond{far} contexts; the shortest, simplest color terms for the target
may also apply to one or both distractors, thus incentivizing the speaker to use more
lengthy descriptions to fully distinguish it. Indeed, even if they \emph{know}
enough simple color terms to distinguish all the colors
lexically, they might be unsure their listeners will and so
resort to modifiers anyway. To assess this hypothesis,
we counted the average number of words and characters per message.
Compared to the baseline \cond{far} context, participants used significantly more words both in the \cond{split} context ($t =  45.85$) and the \cond{close} context ($t = 73.06$). Similar results hold for the character metric.

\paragraph{Comparatives and superlatives}
As noted in \secref{sec:intro}, comparative morphology implicitly
encodes a dependence on the context; a speaker who refers to the
target color as \word{the darker blue} is presupposing that there is
another (lighter) blue in the context. Similarly, superlatives like
\word{the bluest one} or \word{the lightest one} presuppose that all
the colors can be compared along a specific semantic dimension. We
thus expect to see this morphology more often where two or more of the
colors are comparable in this way. To test this, we used the Stanford
CoreNLP part-of-speech tagger \cite{Toutanova2003} to mark the presence or absence of comparatives (JJR or RBR) and superlatives (JJS or RBS) for each message.

We found two related patterns across conditions. First, participants were significantly
more likely to use both comparatives ($z = 37.39$) and superlatives ($z = 31.32$)
when one or more distractors were close to the target. Second, we found evidence of
an asymmetry in  the use of these constructions across the \cond{split} and
\cond{close} contexts. Comparatives were used significantly more often in the
\cond{split} context ($z = 4.4$), where only one distractor was close to the target,
while superlatives were much more likely to be used in the \cond{close} condition
($z = 32.72$).\footnote{We used Helmert coding to test these specific patterns: the first regression coefficient compares the `far' condition to the mean of the other two conditions, and the second regression coefficient compares the `split' condition to the `close' condition.}

\paragraph{Negatives}
In our referential contexts, negation is likely to play a role similar
to that of comparatives: a phrase like \word{not the red or blue one}
singles out the third color, and \word{blue but not bright blue}
achieves a more nuanced kind of comparison. Thus, as with
comparatives, we expect negation to be more likely where one or more
distractors are close to the target. To test this, we counted
occurrences of the string `not' (by far the most frequent negation in
the corpus). Compared to the baseline \cond{far} context, we found that participants were more likely to use negative constructions when one ($z = 27.36$) or both ($z = 34.32$) distractors were close to the target.

\newcommand{\Lex}{\mathcal{L}}
\newcommand{\Costs}{\kappa}
\newcommand{\Messages}{U}
\newcommand{\targetPrior}{P}

\begin{figure*}[t]
  \centering
  \renewcommand{\arraystretch}{0.95}
  \begin{subfigure}[t]{0.23\textwidth}
    \centering
    \begin{tabular}{lr@{\hskip 5pt}r@{\hskip 5pt}r@{}r}
    \toprule
     & \colorContextCompact{3884C7}{02F9FD}{9E6461}{} \\
    \midrule
    blue & \best{1}\p & \best{1}\p & 0\p \\
    teal & 0\p & \best{1}\p & 0\p \\
    dull & \best{1}\p & 0\p & \best{1}\p \\
    \bottomrule
    \end{tabular}
    \caption{The lexicon $\Lex$ defines utterances' truth values.
    Our neural listener skips $\Lex$ and models
    $l_0$'s probability distributions directly.}
    \label{fig:basic-rsa:lex}
  \end{subfigure}
  \hfill
  \begin{subfigure}[t]{0.23\textwidth}
    \centering
    \begin{tabular}{lr@{\hskip 5pt}r@{\hskip 5pt}r@{}r}
    \toprule
     & \colorContextCompact{3884C7}{02F9FD}{9E6461}{} \\
    \midrule
    blue & \best{50} & \best{50}\p & 0\p \\
    teal & 0\p & \best{100}\p & 0\p \\
    dull & \best{50}\p & 0\p & \best{50}\p \\
    \bottomrule
    \end{tabular}
    \caption{The literal listener $l_{0}$ chooses colors compatible with the
             literal semantics of the utterance; other than that, it
             guesses randomly.}
    \label{fig:basic-rsa:l0}
  \end{subfigure}
  \hfill
  \begin{subfigure}[t]{0.23\textwidth}
    \centering
    \begin{tabular}{lr@{\hskip 5pt}r@{\hskip 5pt}r@{}r}
    \toprule
     & \colorContextCompact{3884C7}{02F9FD}{9E6461}{} \\
    \midrule
    blue & \best{50}\p & 33\p & 0\p \\
    teal & 0\p & \best{67}\p & 0\p \\
    dull & \best{50}\p & 0\p & \best{100}\p \\
    \bottomrule
    \end{tabular}
    \caption{The pragmatic speaker $s_{1}$ soft-maximizes the informativity of
             its utterances. (For simplicity, $\alpha = 1$ and
             $\Costs(\utt) = 0$.)}
    \label{fig:basic-rsa:s1}
  \end{subfigure}
  \hfill
  \begin{subfigure}[t]{0.23\textwidth}
    \centering
    \begin{tabular}{lr@{\hskip 5pt}r@{\hskip 5pt}r@{}r}
    \toprule
     & \colorContextCompact{3884C7}{02F9FD}{9E6461}{} \\
    \midrule
    blue & \best{60}\p & 40\p & 0\p \\
    teal & 0\p & \best{100}\p & 0\p \\
    dull & 33\p & 0\p & \best{67}\p \\
    \bottomrule
    \end{tabular}
    \caption{The pragmatic listener $l_{2}$ uses Bayes' rule to infer the target
             using the speaker's utterance as evidence.}
    \label{fig:basic-rsa:l2}
  \end{subfigure}
  \caption{The basic RSA model applied to a reference task (literal semantics and alternative utterances simplified for demonstration). (b)-(d) show conditional probabilities (\%).}
  \label{fig:basic-rsa}
\end{figure*}

\paragraph{WordNet specificity}
We expect speakers to prefer basic color terms wherever they suffice
to achieve the communicative goal, since such terms are most likely to
succeed with the widest range of listeners. Thus, a speaker might
choose \word{blue} even for a clear periwinkle color. However, as the
colors get closer together, the basic terms become too ambiguous, and
thus the risk of specific terms becomes worthwhile (though lengthy
descriptions might be a safer strategy, as discussed above). To
evaluate this idea, we use WordNet \cite{Fellbaum1998} to derive a
specificity hierarchy for color terms, and we hypothesized that
\cond{split} or \cond{close} conditions will tend to lead speakers to go lower in this
hierarchy.

For each message, we transformed adjectives into their closest noun forms (e.g. `reddish' $\rightarrow$ `red'), filtered to include only nouns with `color' in their hypernym paths, calculated the depth of the hypernym path of each color word, and took the maximum depth occurring in a message. For instance, the message ``deep magenta, purple with some pink'' received a score of 9. It has three color terms: ``purple'' and ``pink,'' which have the basic-level depth of 7, and ``magenta,'' which is a highly specific color term with a depth of 9. Finally, because there weren't meaningful differences between words at depths of 8 (``rose'', ``teal'') and 9 (``tan,'' ``taupe''), we conducted our analyses on a binary variable thresholded to distinguish ``high specificity'' messages with a depth greater than 7.
We found a small but reliable increase in the likelihood of ``high specificity'' messages from human speakers in the \cond{split}  ($z = 2.84, p = 0.005$) and \cond{close}  ($z = 2.33, p = 0.02$) contexts, compared to the baseline \cond{far} context.

\section{Models}

We first define the basic RSA model as applied to the
color reference games introduced in \secref{sec:corpus}; a worked example
is shown in \figref{fig:basic-rsa}.

\paragraph{Listener-based listener}
The starting point of RSA is a model of a \term{literal listener}:
\begin{align}
  l_{0}(\target \| \utt, \Lex)
  &\propto
  \Lex(\utt, \target) \targetPrior(\target)
\end{align}
where $\target$ is a color in the context
set $\context$, $\utt$ is a message drawn from a set of possible
utterances $\Messages$, $\targetPrior$ is a prior over colors,
and $\Lex(\utt, \target)$ is a semantic interpretation function that
takes the value $1$ if $\utt$ is true of $\target$, else $0$.
\Figref{fig:basic-rsa:lex} shows the values of $\Lex$ defined for
a very simple context in which $\Messages=\set{\word{blue},
  \word{teal}, \word{dull}}$, and $\context =
\set{\colorContextNarrow{3884C7}{02F9FD}{9E6461}{}}$;
\figref{fig:basic-rsa:l0} shows the corresponding literal listener
$l_0$ if the prior $\targetPrior$ over colors is flat.
(In our scalable extension, we will substitute a neural network model for $l_0$,
bypassing $\Lex$ and allowing for non-binary semantic judgments.)

RSA postulates a model of a \term{pragmatic speaker} (\figref{fig:basic-rsa:s1}) that
behaves according to a distribution that soft-maximizes
a utility function rewarding informativity and penalizing cost:
\begin{align}
  s_{1}(\utt \| \target, \Lex)
  &\propto
  e^{\alpha\log(l_{0}(\target \| \utt, \Lex)) - \Costs(\utt)}
  \label{eq:rsa-s1}
\end{align}
Here, $\Costs$ is a real-valued cost function on utterances, and
$\alpha \in [0,\infty)$ is an inverse temperature parameter governing
the ``rationality'' of the speaker model.
A large $\alpha$ means the pragmatic speaker is expected to choose
the most informative utterance (minus cost)
consistently; a small $\alpha$ means the
speaker is modeled as choosing suboptimal utterances frequently.

Finally, a \term{pragmatic listener} (\figref{fig:basic-rsa:l2})
interprets utterances by reasoning about the behavior of the pragmatic speaker:
\begin{align}
  l_{2}(\target \| \utt, \Lex)
  &\propto
  s_{1}(\utt \| \target, \Lex) \targetPrior(\target)
  \label{eq:rsa-l2}
\end{align}

The $\alpha$ parameter of the speaker indirectly affects the
listener's interpretations: the more reliably the speaker
chooses the optimal utterance for a referent, the more the listener
will take deviations from the optimum as a signal to choose a
different referent.

The most important feature of this model is that the pragmatic
listener $l_{2}$ reasons not about the semantic interpretation
function $\Lex$ directly, but rather about a speaker who reasons about
a listener who reasons about $\Lex$ directly. The back-and-forth
nature of this interpretive process mirrors
that of conversational implicature \cite{Grice75} and
reflects more general ideas from Bayesian cognitive modeling
\cite{Tenenbaum-etal:2011}. The model and its variants have been shown
to capture a wide range of pragmatic phenomena in a cognitively
realistic manner
\cite{Goodman2013,Smith:Goodman:Frank:2013,Kao-etal:2014,Bergen:Levy:Goodman:2014},
and the central Bayesian calculation has proven useful in a variety of
communicative domains
\cite{Tellex2014a,Vogel:Potts:Jurafsky:2013}.

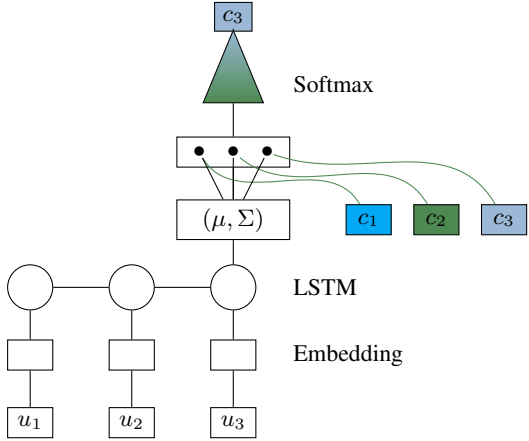
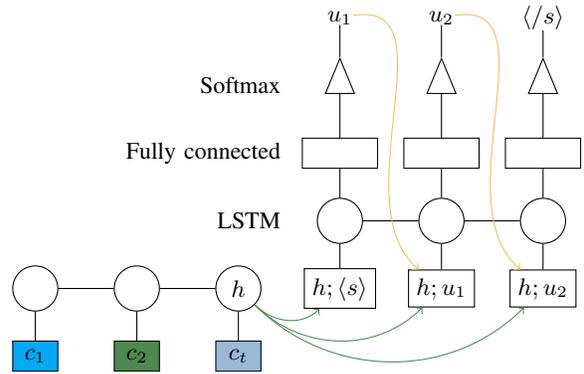
\begin{figure*}[t]
  \centering
   \begin{subfigure}[b]{0.48\textwidth}
    \centering
    \footnotesize
\centering
\begin{tikzpicture}[scale=0.90]

  \tikzstyle{wordvec}=[draw, minimum width=6mm, minimum height=4mm]
  \tikzstyle{embedding}=[draw, minimum width=6mm, minimum height=4mm]
  \tikzstyle{LSTM}=[draw,circle, minimum width=6mm, minimum height=4mm]
  \tikzstyle{linear}=[draw, minimum width=1.5cm, minimum height=4mm]
  \tikzstyle{dist}=[draw, minimum width=1.5cm, minimum height=4mm]
  \tikzstyle{score}=[inner sep=0pt]
  \tikzstyle{softmax}=[draw,
                       isosceles triangle,
                       rotate=90,
                       top color=oursteelblue,
                       bottom color=ourgreen,
                       minimum size=8mm]
  \tikzstyle{colorvec}=[draw, minimum width=6mm, minimum height=4mm]
  \tikzstyle{output}=[draw]

  \tikzstyle{label}=[align=left,text width=2cm]

  \path
  (1,  0) node[wordvec](xx){$u_{1}$}
  (2.5,0) node[wordvec](xy){$u_{2}$}
  (4,  0) node[wordvec](xz){$u_{3}$};

  \path
  (1,  1) node[embedding](hx){}
  (2.5,1) node[embedding](hy){}
  (4,  1) node[embedding](hz){};

  \path
  (1,  2) node[LSTM](lx){}
  (2.5,2) node[LSTM](ly){}
  (4,  2) node[LSTM](lz){};

  \path
  (4, 3) node[linear](linx){$(\mu, \Sigma)$};

  \path
  (6,3) node[colorvec,fill=ourlightblue](cx){$c_{1}$}
  (7,3) node[colorvec,fill=ourgreen](cy){$c_{2}$}
  (8,3) node[colorvec,fill=oursteelblue](cz){$c_{3}$};

  \path
  (4, 4) node[dist](scores){}
  (3.5, 4) node[score](scorex){$\bullet$}
  (4, 4) node[score](scorey){$\bullet$}
  (4.5, 4) node[score](scorez){$\bullet$};

  \path
  (4, 5) node[softmax](sx){};

  \path
  (4, 6) node[output,fill=oursteelblue](yx){$c_{3}$};

  \draw (6, 1) node[label]{Embedding};
  \draw (6, 2) node[label]{LSTM};
  \draw (6, 5) node[label]{Softmax};

  \draw (lx)--(ly)--(lz);
  \draw (xx)--(hx)--(lx);
  \draw (xy)--(hy)--(ly);
  \draw (xz)--(hz)--(lz)--(linx)--(scores)--(sx)--(yx);

  \draw[draw=ourgreen] (cx) to[out=120,in=300] (scorex);
  \draw[draw=ourgreen] (cy) to[out=120,in=320] (scorey);
  \draw[draw=ourgreen] (cz)  to[out=120,in=340] (scorez);

  \draw (linx)--(scorex);
  \draw (linx)--(scorey);
  \draw (linx)--(scorez);

\end{tikzpicture}
    \caption{The $\Listener_{0}$ agent processes tokens $\utt_i$ of a
      color description $\utt$
      sequentially. The final representation is transformed into a
      Gaussian distribution in color space, which is used to score the
      context colors $c_1 \dots c_3$.}
    \label{fig:model:listener}
  \end{subfigure}
  \hfill
  \begin{subfigure}[b]{0.48\textwidth}
    \centering
    \footnotesize
\centering
\begin{tikzpicture}[scale=0.90]

  \tikzstyle{label}=[anchor=east,align=right,text width=2.5cm]
  \tikzstyle{colorvec}=[draw, minimum width=6mm, minimum height=4mm]
  \tikzstyle{hiddencolorvec}=[draw, minimum width=6mm, minimum height=4mm]
  \tikzstyle{token}=[draw]
  \tikzstyle{LSTM}=[draw,circle, minimum width=6mm, minimum height=4mm]
  \tikzstyle{FC}=[draw, minimum width=1cm, minimum height=0.4cm]
  \tikzstyle{softmax}=[draw,isosceles triangle,rotate=90]
  \tikzstyle{output}=[inner sep=0]

  \path
  (1,0) node[colorvec,fill=ourlightblue](cx){$c_{1}$}
  (2.5,0) node[colorvec,fill=ourgreen](cy){$c_{2}$}
  (4,0) node[colorvec,fill=oursteelblue](cz){$c_{t}$};

  \path
  (1, 1) node[LSTM](hx){}
  (2.5, 1) node[LSTM](hy){}
  (4, 1) node[LSTM](hz){$h$};

  \path
  (5.5, 1) node[token](tx){$h;\langle s\rangle$}
  (7, 1) node[token](ty){$h; u_{1}$}
  (8.5,1) node[token](tz){$h; u_{2}$};

  \path
  (5.5, 2) node[LSTM](lx){}
  (7, 2) node[LSTM](ly){}
  (8.5,2) node[LSTM](lz){};

  \path
  (5.5, 3) node[FC](fx){}
  (7, 3) node[FC](fy){}
  (8.5,3) node[FC](fz){};

  \path
  (5.5, 4) node[softmax](sx){}
  (7, 4) node[softmax](sy){}
  (8.5,4) node[softmax](sz){};

  \path
  (5.5, 5) node[output](yx){$u_{1}$}
  (7, 5) node[output](yy){$u_{2}$}
  (8.5,5) node[output](yz){$\langle/s\rangle$};

  \draw (4.75,2) node[label]{LSTM};
  \draw (4.75,3) node[label]{Fully connected};
  \draw (4.75,4) node[label]{Softmax};

  \draw (cx)--(hx);
  \draw (cy)--(hy);
  \draw (cz)--(hz);

  \draw (hx)--(hy)--(hz);
  \draw (lx)--(ly)--(lz);

  \draw (tx)--(lx)--(fx)--(sx)--(yx);
  \draw (ty)--(ly)--(fy)--(sy)--(yy);
  \draw (tz)--(lz)--(fz)--(sz)--(yz);

  \draw[->,draw=ourgreen] (hz) to[out=-45,in=225] (tx);
  \draw[->,draw=ourgreen] (hz) to[out=-45,in=225] (ty);
  \draw[->,draw=ourgreen] (hz) to[out=-45,in=225] (tz);

  \draw[->,draw=ourorange] (yx) to[out=10,in=140] (ty);
  \draw[->,draw=ourorange] (yy) to[out=10,in=140] (tz);
\end{tikzpicture}
    \caption{The $\Speaker_{0}$ agent processes the target color
      $c_{t}$ in context and produces tokens $\utt_i$ of a color description
      sequentially. Each step in production is conditioned by the
      context representation $h$ and the previous word
      produced.}
    \label{fig:model:speaker}
  \end{subfigure}
  \vspace{-2mm}
  \caption{The neural base speaker and listener agents.}
  \label{fig:model}
\end{figure*}

\paragraph{Speaker-based listener}
The definitions of $s_1$ \eqref{eq:rsa-s1} and $l_2$ \eqref{eq:rsa-l2}
give a general method of deriving a speaker from a listener and vice versa.
This suggests an alternative formulation of a pragmatic listener, starting from
a literal speaker:
\begin{align}
  s_{0}(\utt \| \target, \Lex)
  &\propto
  \Lex(\utt, \target) e^{- \Costs(\utt)}
  \label{eq:rsa-s0} \\
  l_{1}(\target \| \utt, \Lex)
  &\propto
  s_{0}(\utt \| \target, \Lex) \targetPrior(\target)
  \label{eq:rsa-l1}
\end{align}
Here, it is the speaker that reasons about the semantics, while the listener
reasons about this speaker.

Both of these versions of RSA pose problems with
scalability, stemming from the set of messages
$\Messages$ and the interpretation function $\Lex$. In most versions of
RSA, these are specified by hand (but see \citealt{Monroe2015}).  This
presents a serious practical obstacle to applying RSA to large data
sets containing realistic utterances. The set $\Messages$
also raises a more fundamental issue: if this set is not finite (as
one would expect from a compositional grammar), then in general there is
no exact way to normalize the $s_{1}$ scores, since the denominator
must sum over all messages. The same problem applies to $s_{0}$, unless
$\Lex$ factorizes in an unrealistically clean way.

Over the next few subsections, we overcome these obstacles by
replacing $l_{0}$ and $s_0$ with RNN-based listener agents,
denoted with
capital letters: $\Listener_0$, $\Speaker_0$.
We use the $\Speaker_0$ agent both as a base model for a pragmatic listener
analogous to $l_1$ in \eqref{eq:rsa-l1} and to acquire sample
utterances for approximating the normalization required in defining
the $s_{1}$ agent in \eqref{eq:rsa-s1}.

\subsection{Base listener}

Our base listener agent $\Listener_0$ (\figref{fig:model:listener}) is an LSTM encoder model that predicts a Gaussian
distribution over colors in a transformed representation space.
The input words are embedded in a 100-dimensional vector space. Word embeddings
are initialized to random normally-distributed vectors ($\mu = 0$, $\sigma = 0.01$)
and trained. The sequence of word vectors is
used as input to an LSTM with 100-dimensional hidden state, and a linear
transformation is applied to the output representation to produce the parameters
$\mu$ and $\Sigma$ of a quadratic form\footnote{The quadratic form is not guaranteed 
to be negative definite and thus define a Gaussian; however, it is for $>95$\% of
inputs. The distribution over context colors is well-defined regardless.}
\[\operatorname{score}(\feat) = -(\feat - \mu)^T \Sigma (\feat - \mu)\]
where $\feat$ is a vector representation of a color. Each color is represented in
its simplest form as a three-dimensional vector in
RGB space. These RGB vectors are then Fourier-transformed as in
\newcite{MonroeGoodmanPotts16_Color} to obtain the representation $\feat$.

The values of $\operatorname{score}(\feat)$ for each of the $\contextlen$
context colors are normalized in log space to
produce a probability distribution over the context colors. We denote this
distribution by $\Listener_0(\target \| \utt, \context; \theta)$, where $\theta$ represents the
vector of parameters that define the trained model.

\subsection{Base speaker}\label{sec:s0}

We also employ an LSTM-based speaker model
$\Speaker_0(\utt \| \target, \context; \phi)$. This speaker serves two purposes:
\begin{enumerate*}[label=\arabic*)]
\item it is used to define a pragmatic listener akin to $l_1$ in \eqref{eq:rsa-l1},
and
\item it provides samples of alternative utterances for each context, to avoid
enumerating the intractably large space of possible utterances.
\end{enumerate*}

The speaker model consists of an LSTM context encoder
and an LSTM description decoder (\figref{fig:model:speaker}). In this model, the colors of the context
$\referent_i \in \context$ are transformed into Fourier representation space,
and the sequence of color representations is passed through an LSTM with
100-dimensional hidden state. The context is reordered to place the target color
last, minimizing the length of dependence between the most important input color
and the output \cite{Sutskever2014} and eliminating the need to represent the
index of the target separately.
The final cell state of this recurrent neural network is concatenated with a
100-dimensional embedding for the previous token output at each step of decoding.
The resulting vector is input along with the previous cell state to the LSTM cell,
and an affine transformation and softmax function are applied to the output to
produce a probability distribution predicting the following token of the description.
The model is substantively similar to well-known models for image caption generation
\cite{Karpathy2015,Vinyals2015}, which use the output of a convolutional neural
network as the representation of an input image and provide this representation
to the RNN as an initial state or first word (we represent the context using
a second RNN and concatenate the context representation onto each input word vector).

\subsection{Pragmatic agents}\label{sec:l2}

Using the above base agents, we define a pragmatic speaker
$\Speaker_{1}$ and a pragmatic listener
$\Listener_{2}$:
\begin{align}
\Speaker_1(\utt \| \target, \context; \theta)
  &= \frac{\Listener_0(\target \| \utt, \context; \theta)^\alpha}{\sum_{\utt'}
    \Listener_0(\target \| \utt', \context; \theta)^\alpha}
    \label{eq:s1} \\
  \Listener_2(\target \| \utt, \context; \theta)
  &=
    \frac{
    \Speaker_1(\utt \| \target, \context; \theta)
    }{
    \sum_{\target'} \Speaker_1(\utt \| \target', \context; \theta)
    }
\end{align}
These definitions mirror those in \eq{eq:rsa-s1} and \eq{eq:rsa-l2}
above, with $\Lex$ replaced by the learned weights $\theta$.

Just as in \eqref{eq:rsa-s1}, the denominator in \eqref{eq:s1} should consist of a sum over
the entire set of potential utterances, which is exponentially large in the
maximum utterance length and might not even be finite.
As mentioned in \secref{sec:s0}, we limit this search by
taking $\numsamples$ samples from $\Speaker_0(\utt \| i, \context; \phi)$ for
each target index $i$, adding the actual utterance from the testing example,
and taking the resulting multiset as the universe of possible utterances,
weighted towards frequently-sampled utterances.\footnote{An alternative would
be to enforce uniqueness within the alternative set, keeping it a true set as in the
basic RSA formulation; this could be done with rejection sampling or beam search
for the highest-scoring speaker utterances. We found that doing so with
rejection sampling hurt model performance somewhat, so we did not
pursue the more complex beam search approach.} Taking a number
of samples from $\Speaker_0$ for each referent in the context gives the pragmatic
listener a variety of informative alternative utterances to consider when
interpreting the true input description.
We have found that $\numsamples$
can be small; in our experiments, it is set to $8$.

To reduce the noise
resulting from the stochastically chosen alternative utterance sets, we also perform
this alternative-set sampling $n$ times and average the resulting probabilities in
the final $\Listener_2$ output. We again choose $n = 8$ as a satisfactory
compromise between effectiveness and computation time.

\paragraph{Blending with a speaker-based agent} A second pragmatic listener
$\Listener_1$ can be formed in a similar way,
analogous to $l_{1}$ in \eqref{eq:rsa-l1}:
\begin{align}
  \Listener_1(\target \| \utt, \context; \phi)
  &=
    \frac{
    \Speaker_0(\utt \| \target, \context; \phi)
    }{
    \sum_{\target'} \Speaker_0(\utt \| \target', \context; \phi)
    }    \label{eq:l1}
\end{align}

We expect $\Listener_1$ to be less accurate than  $\Listener_0$ or
$\Listener_2$, because it is performing
a listener task using only the outputs of a model trained for a speaker task.
However, this difference in training objective can also give the model
strengths that complement those of the two listener-based agents.
One might also expect a realistic model of human language interpretation
to lie somewhere between the ``reflex'' interpretations of the neural base listener and the ``reasoned'' interpretations of
one of the pragmatic models. This has an intuitive justification in people's
uncertainty about whether their interlocutors are speaking pragmatically:
``should I read more into that statement, or take it at face value?''
We therefore also evaluate models defined as
a weighted average of $\Listener_0$ and each of $\Listener_1$ and $\Listener_2$,
as well as an ``ensemble'' model that combines all of these agents.
Specifically, we consider the following blends of neural base models and
pragmatic models, with $\mathbf{L}_{i}$ abbreviating
$\Listener_i(\target \| \utt, \context; \theta, \phi)$ for convenience:
\begin{align}
\mathbf{L}_a &\propto {\mathbf{L}_0}^{\beta_a} \cdot
\mathbf{L}_1^{1-\beta_a}  \label{eq:beta_a} \\
\mathbf{L}_b &\propto {\mathbf{L}_0}^{\beta_b} \cdot
\mathbf{L}_2^{1-\beta_b}  \label{eq:beta_b} \\
\mathbf{L}_e &\propto {\mathbf{L}_a}^{\gamma} \cdot
\mathbf{L}_b^{1-\gamma}  \label{eq:gamma}
\end{align}
The hyperparameters in the exponents allow tuning the
blend of each pair of models---e.g., overriding the neural model with the pragmatic
reasoning in $\Listener_b$. The value of the weights
$\beta_a$, $\beta_b$, and $\gamma$ can be any real number; however, we find that
good values of these weights
lie in the range $[-1, 1]$. As an example, setting $\beta_b = 0$ makes the
blended model $\Listener_b$ equivalent to the pragmatic model $\Listener_2$;
$\beta_b = 1$ ignores the pragmatic reasoning and uses the base model
$\Listener_0$'s outputs; and $\beta_b = -1$
``subtracts'' the base model from the pragmatic model (in log probability space)
to yield a ``hyperpragmatic'' model.

\subsection{Training} \label{sec:training}

We split our corpus into approximately equal train/dev/test sets
(15,665 train trials, 15,670 dev, 15,659 test), ensuring that trials from
the same dyad are present in only one split.
We preprocess the data by
\begin{enumerate*}[label=\arabic*)]
\item lowercasing;
\item tokenizing by splitting off punctuation as well as the endings
\word{\mbox{-er}}, \word{\mbox{-est}}, and
\word{\mbox{-ish}};\footnote{We
only apply this heuristic ending segmentation for the listener; the speaker is trained to produce
words with these endings unsegmented, to avoid segmentation inconsistencies
when passing speaker samples as alternative utterances to the listener.} and
\item replacing tokens that appear once or not at all
in the training split\footnote{1.13\% of training tokens, 1.99\% of dev/test.} with \texttt{<unk>}.
\end{enumerate*}
We also remove
listener utterances and concatenate speaker utterances on the same context.
We leave handling of interactive dialogue to future work (\secref{sec:conclusion}).

We use ADADELTA
\cite{Zeiler2012} and Adam \cite{Kingma2014}, adaptive variants of
stochastic gradient descent (SGD), to train listener and speaker models.
The choice of optimization
algorithm and learning rate for each model were tuned with grid search
on a held-out tuning set consisting of 3,500 contexts.\footnote{For
  $\Listener_0$: ADADELTA, learning rate $\eta = {}$0.2; for
  $\Speaker_0$: Adam, learning rate $\alpha = {}$0.004.}
We also use a fine-grained grid search on this tuning set to determine the
values of the pragmatic reasoning parameters $\alpha$, $\beta$, and $\gamma$.
In our final ensemble $\Listener_e$, we use %
$\alpha = 0.544$, base weights $\beta_a = 0.492$ and $\beta_b = -0.15$, and
a final blending weight $\gamma = 0.491$.
It is noteworthy that the optimal value of $\beta_b$
from grid search is \emph{negative}. The effect of this is to amplify
the difference between $\Listener_0$ and $\Listener_2$: the listener-based
pragmatic model, evidently, is not quite pragmatic enough.

\section{Model results}

\subsection{Speaker behavior}

To compare human behavior with the behavior of our embedded speaker models,
we performed the same behavorial analysis done in \secref{sec:speaker_behavior}.
Results from this analysis are included alongside the human results in
\Tabref{table:metrics}.
Our pragmatic speaker model $\Speaker_1$ did not differ
qualitatively from our base speaker $\Speaker_0$ on any of the metrics,
so we only summarize results for humans and the pragmatic model.

\paragraph{Words and characters} We found human speakers to be more verbose when
colors were closer together, in both number of words and number of characters.
As \tabref{table:metrics} shows, our $\Speaker_{1}$  agent
shows the same increase in utterance length in the \cond{split} ($t = 18.07$) and \cond{close} ($t = 35.77$) contexts compared to the \cond{far} contexts.

\paragraph{Comparatives and superlatives} Humans used more comparatives and
superlatives when colors were closer together; however, comparatives were
preferred in the  \cond{split} contexts, superlatives in
the \cond{close}
 contexts. Our pragmatic speaker shows the first of these two patterns,
producing  more comparatives ($z = 14.45$) and superlatives ($z = 16$) in the \cond{split} or \cond{close} conditions than in the baseline \cond{far} condition. It does not, however, capture the peak in comparative use in the \emph{split} condition. This suggests that our model is simulating the human strategy at some level, but that more subtle patterns require further attention.

\paragraph{Negations} Humans used more negations when the colors were closer
together. Our pragmatic speaker's use of negation shows the  same relationship to the context ($z = 8.55$ and $z= 16.61$, respectively).

\paragraph{WordNet specificity} Humans used more ``high specificity'' words
(by WordNet hypernymy depth) when the colors were closer together.
Our pragmatic speaker showed a similar effect ($z = 2.65, p =0.008$ and
$z = 2.1, p =0.036$, respectively).

\begin{table}[t]
\centering
\begin{tabular}{lrr}
  \toprule
  model & accuracy (\%) & perplexity \\
  \midrule
  $\Listener_0$                                      & 83.30 & 1.73 \\
  $\Listener_1 = \Listener(\Speaker_0)$              & 80.51 & 1.59 \\
  $\Listener_2 = \Listener(\Speaker(\Listener_0))$   & 83.95 & 1.51 \\
  $\Listener_a = \Listener_0 \cdot \Listener_1$      & 84.72 & 1.47 \\
  $\Listener_b = \Listener_0 \cdot \Listener_2$      & 83.98 & 1.50 \\
  $\Listener_e = \Listener_a \cdot \Listener_b$      & \best{84.84} & \best{1.45}
  \\[0.5ex]
  human & \oracle{90.40} \\
  \midrule
  $\Listener_0$                                      & 85.08 & 1.62 \\
  $\Listener_e$                                      & \best{86.98} & \best{1.39}
  \\[0.5ex]
  human & \oracle{91.08} \\
  \bottomrule
\end{tabular}
\caption{Accuracy and perplexity of the base and pragmatic listeners and
various blends (weighted averages, denoted $A \cdot B$).
Top: dev set; bottom: test set.}
\label{table:modelAccuracy}
\end{table}

\begin{figure}
\centering
\includegraphics[scale = .5]{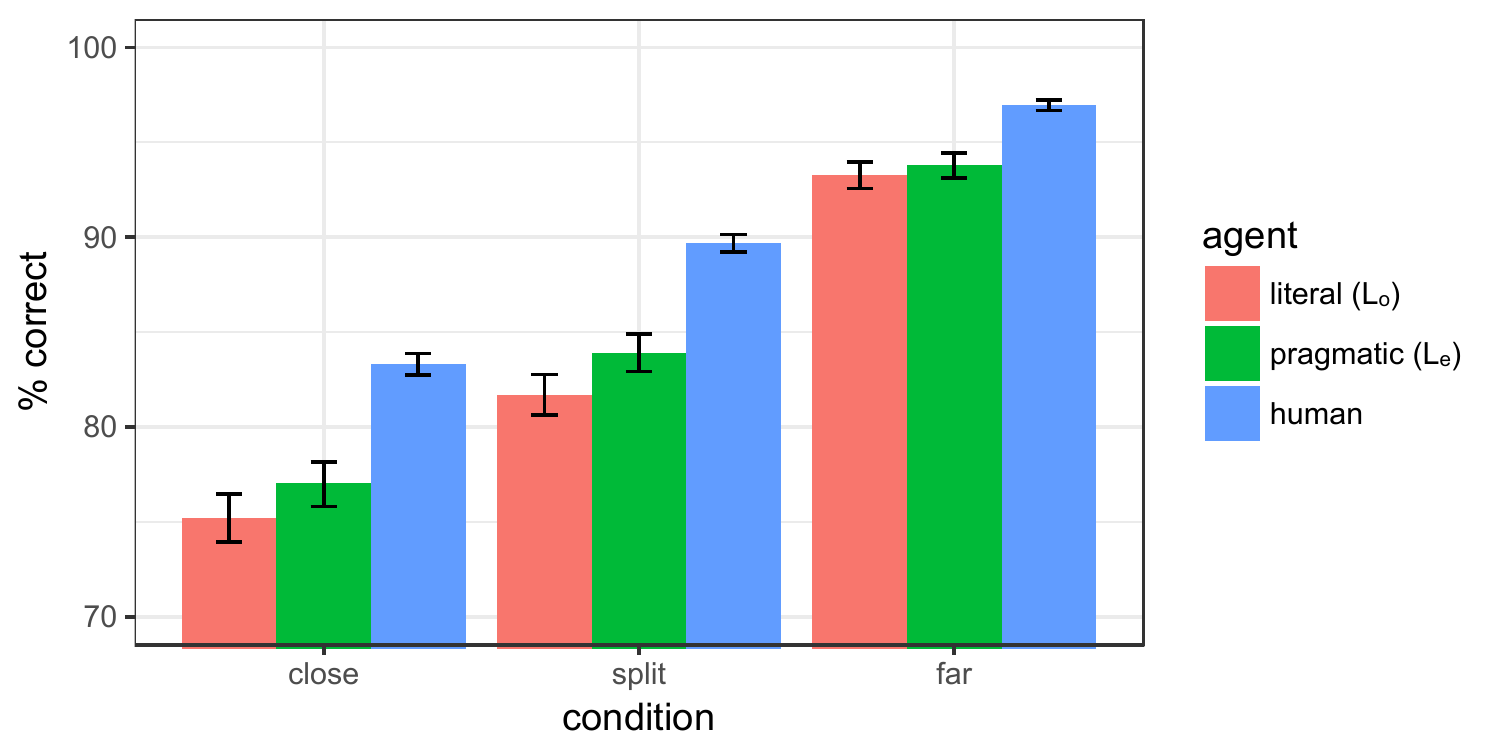}

\includegraphics[scale = .5]{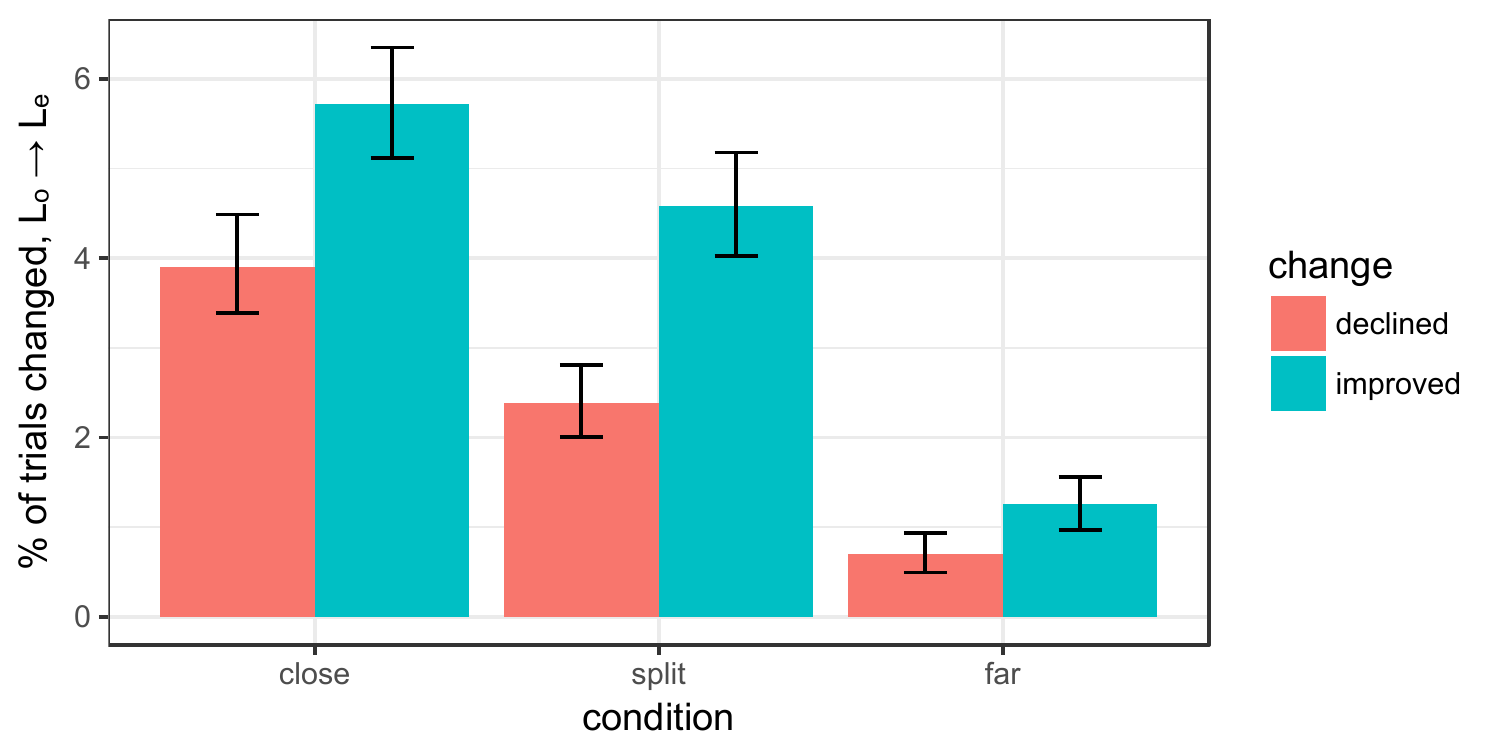}
\caption{Human and model reference game performance (top) and fraction of examples improved and
declined from $\Listener_0$ to $\Listener_e$ (bottom) on the dev set, by condition.}
\label{fig:listenerAccuracy}
\end{figure}

\subsection{Listener accuracy}

\newcommand{\intended}{\framebox}
\newcommand{\speakerpct}{\textit}

\begin{figure*}[t!]
\centering
\begin{tabular}{lr@{\hskip 5pt}r@{\hskip 5pt}r@{}r}
    \toprule
    $\Listener_0$ & \colorContext{3884C7}{02F9FD}{9E6461}{} \\
    \midrule
    \textbf{blue} &             9 & \best{   91} & $<$1
    \\[1ex]
    true blue     & \intended{\gz{}11} & \best{     89} & $<$1 \\
    light blue    & \intended{\zz{}$<$1} & \best{$>$99} & $<$1 \\
    brightest     & $<$1 & \intended{\best{\z{}$>$99}} & $<$1 \\
    bright blue   & $<$1 & \intended{\best{\z{}$>$99}} & $<$1 \\
    red           & $<$1 &     1 & \intended{\best{\gz{}99}} \\
    purple        & $<$1 &     2 & \intended{\best{\gz{}98}} \\
    \midrule
    $\Speaker_1$ & \colorContext{3884C7}{02F9FD}{9E6461}{}  \\
    \midrule
    \textbf{blue} &             41  &    19 & $<$1
    \\[1ex]
    true blue     & \intended{\best{\gz{}47}} &    19 & $<$1 \\
    light blue    & \intended{\gzz{}5} & \best{   20} & $<$1 \\
    brightest     & $<$1 & \intended{\best{\gz{}20}} & $<$1 \\
    bright blue   &    2 & \intended{\best{\gz{}20}} & $<$1 \\
    red           &    1 &     2 & \intended{\best{\gz{}50}} \\
    purple        &    5 &     1 & \intended{\best{\gz{}50}} \\
    \midrule
    $\Listener_2$ & \colorContext{3884C7}{02F9FD}{9E6461}{} \\
    \midrule
    \textbf{blue} & \best{            68}  &    32 & $<$1 \\
    \midrule
    $\Speaker_0$   & \speakerpct{5.71}  & \speakerpct{\best{7.63}} & \speakerpct{0.01} \\
    $\Listener_1$  & 43  & \best{57} & $<$1 \\
    \midrule
    $\Listener_a$  & \best{50} & \best{50} & $<$1 \\
    $\Listener_b$  & \best{68} & 32 & $<$1 \\
    $\Listener_e$  & \best{59} & 41 & $<$1 \\
   \bottomrule
\end{tabular}
\qquad
\begin{tabular}{lr@{\hskip 5pt}r@{\hskip 5pt}r@{}r}
    \toprule
    $\Listener_0$ & \colorContext{718E82}{62909D}{AAC639}{} \\
    \midrule
    \textbf{drab green not the bluer one} & $<$1 & $<$1 & \best{$>$99}
    \\[1ex]
    gray            & \intended{\gz{}\best{96}} &  4 & $<$1 \\
    blue dull green & \intended{\gz{}24} & \best{76} & $<$1 \\
    blue            &  $<$1 & \intended{\best{\z{}$>$99}} &   $<$1  \\
    bluish          &  $<$1 & \intended{\best{\z{}$>$99}} &   $<$1  \\
    green           & 4 & 1 & \intended{\gz{}\best{95}} \\
    yellow          & $<$1 & $<$1 & \intended{\best{\z{}$>$99}} \\
    \midrule
    $\Speaker_1$ & \colorContext{718E82}{62909D}{AAC639}{}  \\
    \midrule
    \textbf{drab green not the bluer one} & 1  & $<$1 & \best{34}
    \\[1ex]
    gray            & \intended{\gz{}\best{58}} & 5 & $<$1 \\
    blue dull green & \intended{\gz{}27} & 28 & $<$1 \\
    blue            & 2 & \intended{\best{\gz{}32}} &   $<$1  \\
    bluish          & 1 & \intended{\best{\gz{}32}} &   $<$1  \\
    green           & 10 & 3 & \intended{\gz{33}} \\
    yellow          & $<$1 & $<$1 & \intended{\gz{}\best{34}} \\
    \midrule
    $\Listener_2$  & \colorContext{718E82}{62909D}{AAC639}{} \\
    \midrule
    \textbf{drab green not the bluer one} & 5  & $<$1 & \best{95} \\
    \midrule
    $\Speaker_0$ ($\times 10^{-9}$)   & \speakerpct{\best{5.85}} & \speakerpct{0.38} & \speakerpct{$<$0.01} \\
    $\Listener_1$  & \best{94} & 6 & $<$1 \\
    \midrule
    $\Listener_a$  & \best{92} & 6 & 2 \\
    $\Listener_b$  & 8 & 1 & \best{91} \\
    $\Listener_e$  & \best{63} & 6 & 32 \\
   \bottomrule
\end{tabular}
\caption{Conditional probabilities (\%) of all agents
for two dev set examples. The target color is boxed, and the human
utterances (\word{blue}, \word{drab green not the bluer one}) are \textbf{bolded}.
Boxed cells for
alternative utterances indicate the intended target;
largest probabilities are in \textbf{bold}.
$\Speaker_0$ probabilities (\speakerpct{italics}) are
normalized across all utterances.
Sample sizes are reduced to save space; here,
$m = 2$ and $n = 1$ (see \secref{sec:l2}).}
\label{fig:rsaExample}
\end{figure*}

\Tabref{table:modelAccuracy} shows the accuracy
and perplexity of the base listener $\Listener_0$, the pragmatic listeners
$\Listener_1$ and $\Listener_2$, and the blended models $\Listener_a$,
$\Listener_b$, and $\Listener_e$ at resolving the human-written color
references. Accuracy differences are significant\footnote{$p <{}$0.012, approximate
permutation test \cite{Pado2006} with Bonferroni correction, 10,000 samples.}
for all pairs except $\Listener_2$/$\Listener_b$ and $\Listener_a$/$\Listener_e$.
As we expected, the speaker-based $\Listener_1$ alone
performs the worst of all the models. However, blending it with $\Listener_0$
doesn't drag down $\Listener_0$'s performance but rather produces a considerable
improvement compared to both of the original models, consistent with our
expectation that the listener-based and speaker-based models have complementary
strengths.

\begin{figure}[t]
\centering
\includegraphics[trim={1cm 0.8cm 2.1cm 0},clip,width = \columnwidth]{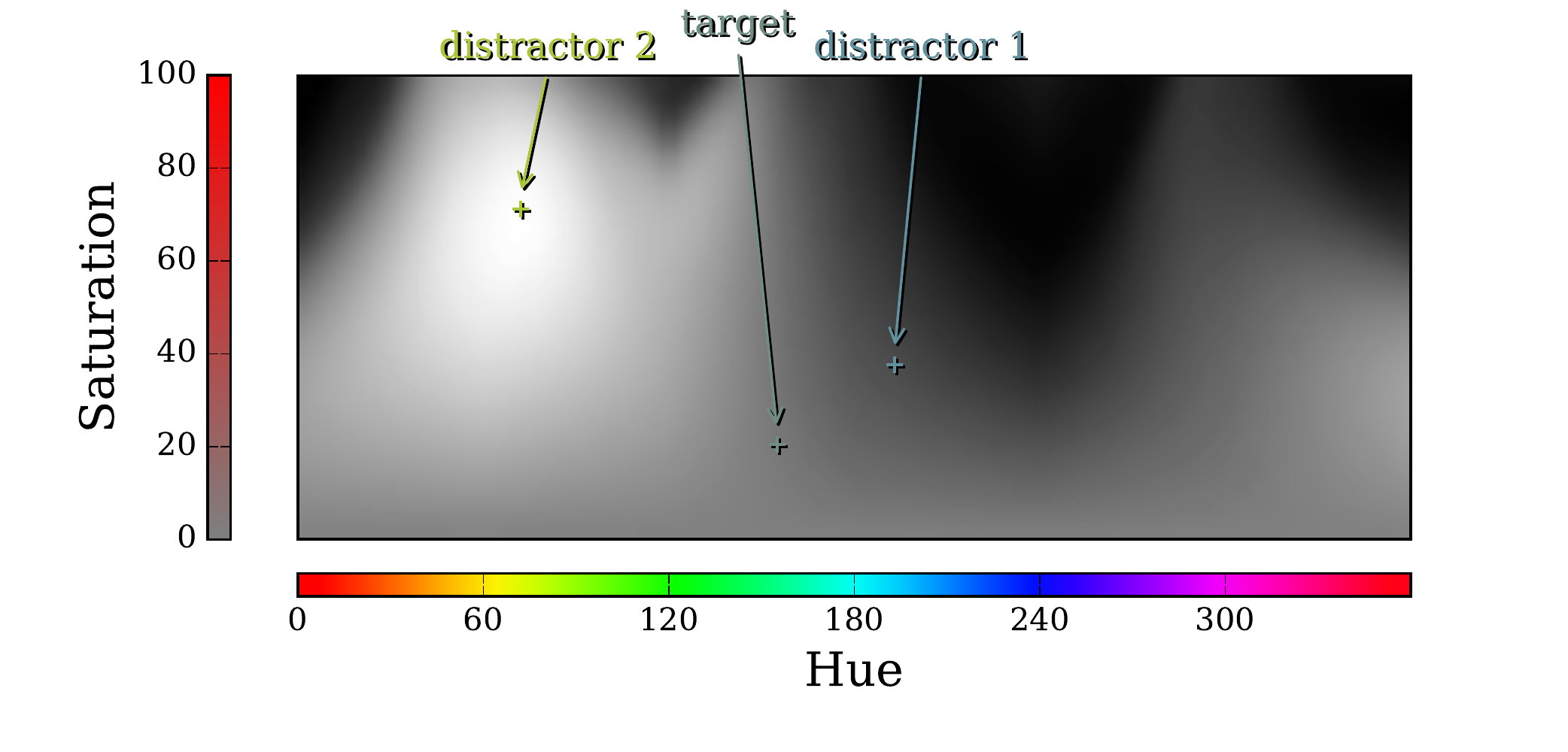}
\caption{$\Listener_0$'s log marginal probability density, marginalizing over
V (value) in HSV space, of color conditioned on the utterance
\word{drab green not the bluer one}. White regions have higher probability.
Labeled colors are the three colors from the right column of
\figref{fig:rsaExample}. }
\label{fig:gaussian}
\end{figure}

We observe that
$\Listener_2$ significantly outperforms its own base model $\Listener_0$,
showing that pragmatic reasoning on its own contributes positively. Blending the
pragmatic models with the base listener also improves over both individually,
although not significantly in the case of $\Listener_b$ over $\Listener_2$.
Finally, the
most effective listener combines both pragmatic models with the base listener.
Plotting the number of examples changed by condition on the dev set
(\figref{fig:listenerAccuracy}) reveals that
the primary gain from including the pragmatic models is in the
\cond{close} and \cond{split} conditions, when the
model has to distinguish highly similar colors and
often cannot rely only on basic color terms.
On the test set, the final ensemble improves
significantly\footnote{$p <{}$0.001, approximate
permutation test, 10,000 samples.} over the base model on both metrics.

\section{Model analysis}

Examining the full probability tables for various dev set examples offers insight
into the value of each model in isolation and how they complement each other when
blended together. In particular, we see that the listener-based ($\Listener_2$)
and speaker-based ($\Listener_1$) pragmatic listeners each overcome a different
kind of ``blind spot'' in the neural base listener's understanding ability.

First, we inspect examples in which $\Listener_2$ is superior to $\Listener_0$.
In most of these examples, the alternative utterances sampled from $\Speaker_0$
for one of the referents $i$ fail to identify
their intended referent to $\Listener_0$. The pragmatic listener interprets
this to mean that referent $i$ is inherently difficult to refer to,
and it compensates by increasing referent $i$'s probability.

This is beneficial when $i$ is the true target. The left column of 
\figref{fig:rsaExample} shows one such example:
a context consisting of a somewhat prototypical blue,
a bright cyan, and a purple-tinged brown, with the utterance \textit{blue}. The
base listener interprets this as referring to the cyan with 91\% probability,
perhaps due to the extreme saturation of the cyan maximally activating certain
parts of the neural network. However, when the pragmatic model takes samples
from $\Speaker_0$ to probe the space of alternative utterances, it becomes
apparent that indicating the more ordinary blue to the listener is difficult:
for the utterances chosen by $\Speaker_0$ intending this referent (\textit{true blue},
\textit{light blue}), the listener also
chooses the cyan with $>$89\% confidence.

Pragmatic reasoning overcomes this difficulty. Only two utterances in the
alternative set (the actual utterance \textit{blue} and the sampled alternative
\textit{true blue}) result in any appreciable probability mass on the true target,
so the pragmatic listener's model of the speaker predicts that the speaker
would usually choose one of these two utterances for the prototypical blue.
However, if the target
were the cyan, the speaker would have many good options. Therefore, the
fact that the speaker chose \textit{blue} is interpreted as evidence for the
true target. This mirrors the back-and-forth reasoning behind the
definition of conversational implicature \cite{Grice75}.

This reasoning can be harmful when $i$ is one of the distractors: the pragmatic
listener is then in danger of overweighting the distractor and incorrectly
choosing it. This is a likely reason for the small performance difference
between $\Listener_0$ and $\Listener_2$. Still, the fact that $\Listener_2$
is more accurate overall, in addition to the negative value of $\beta_b$
discovered in grid search, suggests that the pragmatic reasoning provides
value on its own.

However, the final performance improves greatly when we incorporate both
listener-based and speaker-based agents. To explain this improvement, we examine
examples in which both listener-based agents $\Listener_0$ and
$\Listener_2$ give the wrong answer but are overridden by the speaker-based
$\Listener_1$ to produce the correct referent. The discrepancy between the two
kinds of models in many of these examples can be explained by the fact that the
speaker takes the context as input, while the listener does not. The listener is
thus asked to predict a region of color space from the utterance \term{a priori},
while the speaker can take into account relationships between the context colors
in scoring utterances.

The right column of \figref{fig:rsaExample} shows an example of this. The context
contains a grayish green (the target), a grayish blue-green (``distractor 1''),
and a yellowish green (``distractor 2'').
The utterance from the human speaker is \word{drab green not the bluer one},
presumably intending \word{drab} to exclude the brighter yellowish green. However,
the $\Listener_0$ listener must choose a region of color space to predict based on
the utterance alone, without seeing the other context colors.

\Figref{fig:gaussian} shows a visualization of the listener's prediction.
The figure is a heatmap of the probability density output by the listener,
as a function of hue and saturation in HSV (hue, saturation, value) space.
We use HSV here, rather than the RGB coordinate system used by the model,
because the semantic constraints are more clearly expressed in terms of
hue and saturation components: the color should be \word{drab} (low-saturation)
and \word{green} (near 120 on the hue spectrum) but not \word{blue}
(near 240 in hue). The utterance does not constrain the value
(roughly, brightness--darkness) component, so we sum over this component to
summarize the 3-dimensional distribution in 2 dimensions.

The $\Listener_0$ model correctly interprets all of these constraints:
it gives higher probability to low-saturation colors and greens, while avoiding
bluer colors. However, the result is a
probability distribution nearly centered at distractor 2, the brighter green.
In fact, if we were not comparing it to the other colors in the context,
distractor 2 would be a very good example of a drab green that is not bluish.

The speaker $\Speaker_0$, however, produces utterances
conditioned on the context; it has successfully learned that \textit{drab}
would be more likely as a description of the grayish green than as a description
of the yellowish one in this context. The speaker-based listener $\Listener_1$
therefore predicts the true target, with greater confidence than $\Listener_0$
or $\Listener_2$. This prediction results in the blends $\Listener_a$ and
$\Listener_e$ preferring the true target, allowing the speaker's perspective
to override the listener's.

\section{Related work}

Prior work combining machine learning with probabilistic pragmatic reasoning
models has largely focused on the speaker side, i.e., generation.
\newcite{Golland2010} develop a pragmatic speaker model,
$\Speaker(\Listener_0)$, that reasons about log-linear listeners trained on human
utterances containing spatial references in virtual-world environments.
\newcite{Tellex2014a} apply a similar technique, under the name
\term{inverse semantics}, to create a robot that can informatively ask
humans for assistance in accomplishing tasks.
\newcite{Meo2014} evaluate a model of color description generation
\cite{McMahan2015} on the color
reference data of \newcite{Baumgaertner2012} by creating an $\Listener(\Speaker_0)$
listener.
\newcite{Monroe2015} implement
an end-to-end trained $\Speaker(\Listener(\Speaker_0))$ model for referring
expression generation in a reference game task. Many of these models require
enumerating the set of possible utterances for each context, which is infeasible when
utterances are as varied as those in our dataset.

The closest work to ours that we are aware of is that of
\newcite{AndreasKlein16_NeuralPragmatics}, who also combine neural speaker
and listener models in a reference game setting. They propose a
pragmatic speaker, $\Speaker(\Listener_0)$, sampling from a neural
$\Speaker_0$ model to limit the search space and regularize the model toward
human-like utterances. We show these techniques help in
listener (understanding) tasks as well. Approaching pragmatics from the listener
side requires either inverting the pragmatic reasoning (i.e., deriving a
listener from a speaker), or adding another step of recursive reasoning,
yielding a two-level derived pragmatic model
$\Listener(\Speaker(\Listener_0))$. We show both approaches contribute
to an effective listener.

\section{Conclusion} \label{sec:conclusion}

In this paper, we present a newly-collected corpus of color descriptions from
reference games, and we show that a pragmatic reasoning agent
incorporating neural listener and speaker models
interprets color descriptions in context better than the listener alone.

The separation of referent and utterance representation in our base speaker and
listener models in principle allows easy substitution of referents other than colors
(for example, images), although the performance of the listener agents could be
limited by the representation of utterance semantics as a Gaussian distribution in
referent representation space. Our pragmatic agents also rely on the ability to 
enumerate the set of possible referents. Avoiding this enumeration, as would be 
necessary in tasks with intractably large referent spaces, is a challenging 
theoretical problem for RSA-like models.

Another important next step is to pursue
multi-turn dialogue. As noted in \secref{sec:corpus},
both participants in our reference game task could use the chat window
at any point, and more than half of dyads
had at least one two-way interaction. Dialogue agents are more
challenging to model than isolated speakers and listeners,
requiring long-term planning, remembering previous utterances, and
(for the listener) deciding when to ask for clarification or commit to a referent
\cite{Lewis79_Scorekeeping,BrownYule83_Discourse,Clark96,Roberts96_InformationStructureDiscourse}.
We release our
dataset\footnote{\label{foot:release}\url{https://cocolab.stanford.edu/datasets/colors.html}}
with the expectation that others may find interest in these challenges as well.

\section*{Acknowledgments}

We thank Kai Sheng Tai and Ashwin Paranjape for helpful feedback.
This material is based in part upon work supported by the 
Stanford Data Science Initiative and by the NSF
under Grant No.\ BCS-1456077.
RXDH was supported by the Stanford Graduate Fellowship and the
NSF Graduate Research Fellowship under Grant No.\ DGE-114747.
NDG was supported by the Alfred P.\ Sloan Foundation Fellowship and
DARPA under Agreement No.\ FA8750-14-2-0009.
Any opinions, findings, and conclusions or recommendations 
expressed in this material are those of the authors and do not necessarily
reflect the views of the NSF, DARPA, or the Sloan Foundation.

\bibliography{colors}
\bibliographystyle{acl2012}

\end{document}